\documentclass[10pt,twocolumn,letterpaper]{article}

\usepackage{iccv}
\usepackage{times}
\usepackage{epsfig}
\usepackage{graphicx}
\usepackage{amsmath}
\usepackage{amssymb}

\usepackage{algorithmic}
\usepackage{algorithm}

\usepackage[breaklinks=true,bookmarks=false]{hyperref}

\iccvfinalcopy % *** Uncomment this line for the final submission

\def\iccvPaperID{6} % *** Enter the ICCV Paper ID here

\ificcvfinal\fi

\begin{document}

%%%%%%%%% TITLE
\title{End-to-End Video Captioning}

\author{Silvio Olivastri$^{1}$, Gurkirt Singh$^{2}$ and Fabio Cuzzolin$^{2}$\\
$^{1}$AI Labs, Bologna, Italy 
$^{2}$Oxford Brookes University, United Kingdom\\
{\tt\small silvio@ailabs.it}, {\tt\small  gurkirt.singh-2105@brookes.ac.uk}, {\tt\small fabio.cuzzolin@brookes.ac.uk}
}

\maketitle
\ificcvfinal\thispagestyle{empty}\fi

\begin{abstract}  
Building correspondences across different modalities, such as video and language, has recently become critical in many visual recognition applications, such as video captioning. Inspired by machine translation, recent models tackle this task using an encoder-decoder strategy. The (video) encoder is traditionally a Convolutional Neural Network (CNN), while the decoding (for language generation) is done using a Recurrent Neural Network (RNN). Current state-of-the-art methods, however, train encoder and decoder separately. CNNs are pretrained on object and/or action recognition tasks and used to encode video-level features. The decoder is then optimised on such static features to generate the video's description. This disjoint setup is arguably sub-optimal for input (video) to output (description) mapping. 

In this work, we propose to optimise both encoder and decoder simultaneously in an end-to-end fashion. In a two-stage training setting, we first initialise our architecture using pre-trained encoders and decoders -- then, the entire network is trained end-to-end in a fine-tuning stage to learn the most relevant features for video caption generation. In our experiments, we use GoogLeNet and Inception-ResNet-v2 as encoders and an original Soft-Attention (SA-) LSTM as a decoder. Analogously to gains observed in other computer vision problems, we show that end-to-end training significantly improves over the traditional, disjoint training process. We evaluate our End-to-End (EtENet) Networks on the Microsoft Research Video Description (MSVD) and the MSR Video to Text (MSR-VTT) benchmark datasets, showing how EtENet achieves state-of-the-art performance across the board.
\end{abstract}
\section{Introduction}

% Intro of video captioning problem
\textbf{Video captioning} is the problem of generating textual descriptions based on video content, a key functionality to pave the way for, e.g., talking cars, surgical robots or factories. 
%Some of its exciting applications include human-robot interaction, automated video content description and assisting the visually impaired by describing the content of movies to them. 
The task is particularly challenging for approaches should capture not only the objects, scenes, and activities present in the input video (i.e. address %tasks such as 
video tagging, object and action recognition), %and object recognition), 
but also express how these objects, scenes, and activities relate to each other in a spatial and temporal fashion using a natural language construction.

% Traditional  methods and their limitations
%As shown in Wu~\etal~\cite{Wu}, 
Two major approaches to video captioning exist~\cite{Wu}: 
template-based language models and sequence learning-based ones.
The former class of methods detect words from the visual content (e.g. via object detection) to then generate the desired sentence using grammatical constraints such presence of a subject, 
verbs, object triplets, and so on. Interesting studies in this sense were conducted in~\cite{Guadarrama, Rohrbach, XuR}. 
The latter group of approaches, instead, learn a probability distribution from %the common space (
a set of feature vectors extracted from the video %) of visual content 
to flexibly generate a sentence without using any specific %pattern or 
language template.
Examples of this second category of approaches are~\cite{Venugopalan,Yao,Wang}.

\textbf{Encoder-decoder frameworks}. Thanks to the recent developments of deep learning frameworks such as Long Short-Term Memory (LSTM)~\cite{Hochreiter} networks and Gated Recurrent Units (GRU)~\cite{Cho}, as well as of machine translation techniques such as~\cite{Sutskever}, the currently dominant approach to video captioning is based on sequence learning in an \emph{encoder-decoder} framework. 
In this setting, the encoder represents the input video sequence as a fixed-dimension feature vector, which is then fed to the decoder to generate the output sentence one word at a time. At each time step in the decoder, the current input (the word) and the previously generated hidden states of the output sequence are used to predict the next word and the hidden state. 

One of the most severe drawbacks of such models, however, is that the underlying video content feature space is static and does not change during the training process. More specifically, an encoder (typically, a Convolutional Neural Network, CNN) is pre-trained on datasets designed for different tasks, to be then used as feature extractor for video-captioning. Although this makes some sense given the multi-task nature of the video captioning process illustrated above, the resulting disjoint training process, in which the decoder is trained on the captioning task with static features as input, is inherently suboptimal. 
Recent state-of-the-art methods~\cite{Yao, Pan2, Zhang} try to address this issue by capturing dynamic temporal correspondences between feature vectors corresponding to different video frames. However, these works do not address the basic fact that video captioning may well require the system to learn task-specific features that will necessarily differ from those learned for the  action classification, video tagging or object detection tasks for which the CNN encoder was trained.

\begin{figure}[t!]
\begin{center}
%\fbox{\includegraphics[width=0.85\linewidth]{img/fig1.png}}
\includegraphics[width=1.0\linewidth]{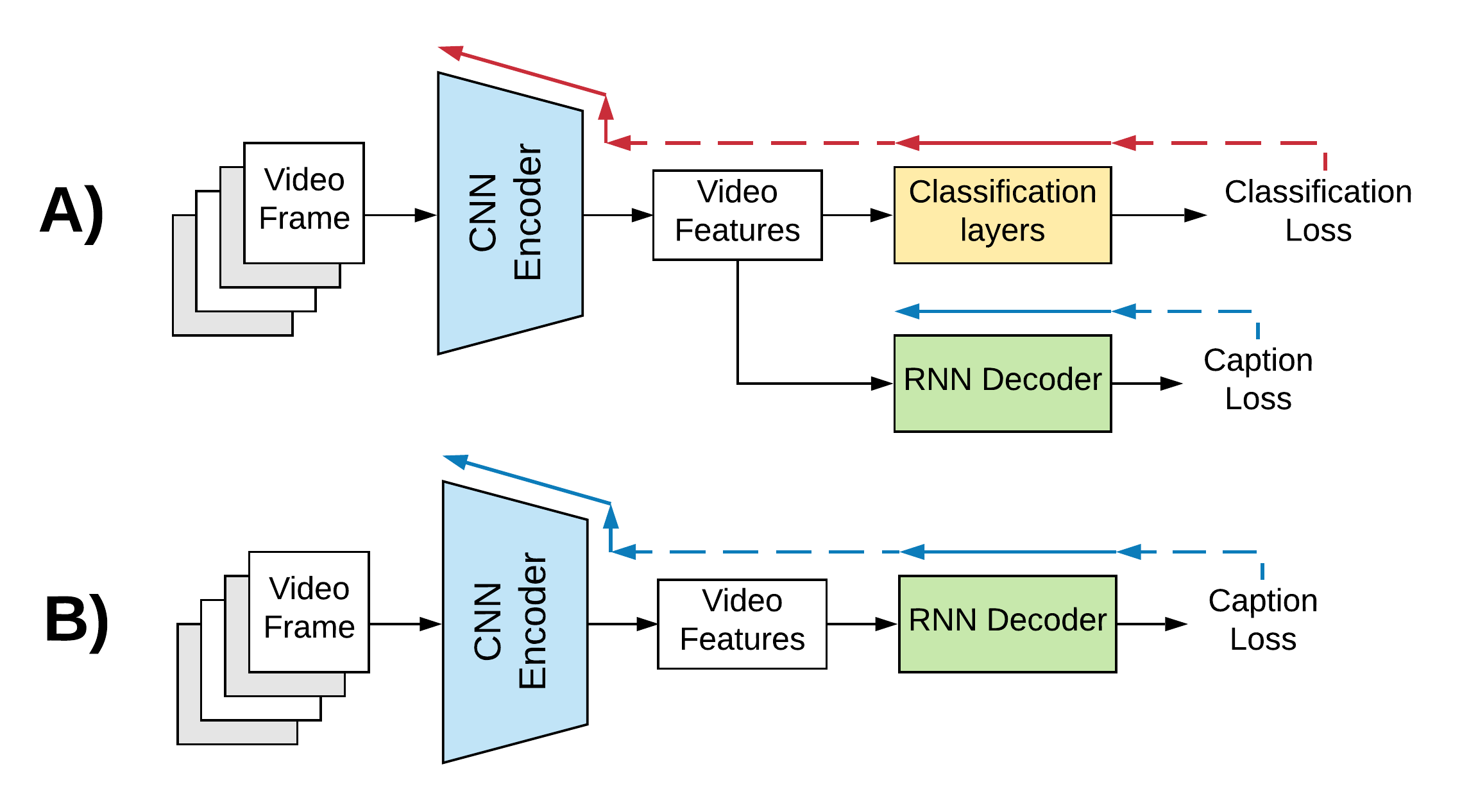}
\end{center}
\caption{Gradient flow comparison between the disjoint training of the RNN decoder and the CNN encoder (A) and the end-to-end training of both encoder and decoder (B). 
In the first case (A), the CNN encoder does not update its parameters in dependence of the captioning loss, but as a function of just the classification loss. Only the decoder updates its parameters as a function of the gradient of the captioning loss (blue arrow).
In the end-to-end case considered here (B), the parameters of both CNN and RNN are updated according to the evolution of the gradients from the captioning loss. Gradient flow is again depicted by a blue arrow.} \vspace{-6mm}
\label{fig:gradientflow}
\end{figure}

\textbf{From complex decoders to learning task-specific features}.
Our view is that \emph{a decoder designed for video captioning should instead be able to learn from task-specific features}.
As shown in Figure \ref{fig:gradientflow} (A) in all previews architectures the encoder is never trained, (e.g., the gradient is not updated during the learning of the encoder part). In addition, the CNN implementing the encoding is trained using a different loss function aimed at solving a different task. 

%Starting from that fact, we realize that 
In order to compensate for this fundamental flaw, 
new decoding mechanisms must be implemented to learn better video features.
As a result decoders (and the associated training procedures) have consistently increased in complexity over the years. To illustrate this point: Venugopalan \etal~\cite{Venugopalan} (2015) used a vanilla LSTM; Yu \etal~\cite{Yu} (2016) multiple Gate Recurrent Units with two attention mechanisms; Pan \etal~\cite{Pan} (2017) implemented a Transfer Unit in combination with two LSTMs; finally, PickNet \cite{ChenY} (2018) achieved excellent results using reinforcement learning to train a deep net able to identify the most relevant frames.

\emph{Our approach is in radical contrast with these previous attempts}.
Rather than designing new, expensive decoding mechanisms to learn better video features,
we force the encoder's feature extractor to focus its attention on 'significant' objects in order to generate good textual descriptions,
exploiting the ability of convolutional networks to
%CNNs are known to be able to 
extract relevant features from images or videos. %thus, rather than create yet another type of decoder, 

\textbf{Our proposal: end-to-end training}.
We propose to address this problem by bringing forward the end-to-end training of both encoder and decoder as a whole, as shown in Figure~\ref{fig:gradientflow} (B).
Our philosophy is inspired by the success of end-to-end trainable network architectures in image recognition~\cite{Krizhevsky}, image segmentation~\cite{long2015fully}, object detection~\cite{Ren} and image-captioning~\cite{Lu,Vinyals}, but \emph{was never before adopted in a video captioning setting}.

\iffalse
As a significant example, in video generation the TGANs-C model~\cite{Pan3} is trained end-to-end by optimising three different kinds of losses: a video-level and a frame-level matching-aware loss to correct 
%the label of real or synthetic videos/frames 
video and frame labels, and to align video/frames with the most correct caption, respectively; and a temporal coherence loss to emphasise temporal consistency. In image captioning, Vinyals \etal~\cite{Vinyals} show how training the whole model achieves superior performance using just the negative log-likelihood loss. Famously, Faster R-CNN~\cite{Ren} region proposal-based CNN architectures achieved what was then state-of-the-art object detection accuracy at the PASCAL VOC challenges in both 2007 and 2012. 
\fi

The reason is that, 
%as evidence suggests, 
if done naively, training end-to-end the resulting large scale, heterogeneous network has 
%such a large computational cost to become
a prohibitive computational cost, especially on machines with a limited number of GPUs.
Introducing the technique into video captioning, however, is quite imperative for end-to-end training allows for simple inference~\cite{Vinyals} and can handle complex, multi-task problems best described by multiple losses~\cite{Pan3}.

\textbf{Two-stage learning for efficient end-to-end training}.
An efficient training procedure is then crucial to unlock the potential of end-to-end training for video captioning.

In this paper we address this issue in two ways. 
\\
Firstly, we propose a \emph{two-stage learning process}, in which encoder and decoder are first trained separately, after initialisation from disjoint models, to leave fine tuning to be conducted in the second stage in a fully end-to-end fashion.
In second place, during training we accumulate gradients over multiple steps in order to update parameters only after the required effective batch size is achieved. 
This leads to an increased number of iterations, which is however mitigated by our two-stage approach.
%This approach is slower to train as compared to the separate training of the two components, because of the increase in the number of iterations required.
%Such a process is simple, and can be implemented in current deep learning platforms by just two lines of code.
\\
%\textbf{A modified Soft-Attention decoder}.
Additionally, to further improve performance, we propose significant changes to the Soft-Attention decoder by Yao \etal~\cite{Yao}, while preserving its simplicity of concept. 
%Starting from that baseline we improve it significantly taking care of leave the approach very simply like the original one. 
The major improvements we propose are presented in Section \ref{subsec:decoder}, and encompass structural architectural changes, a different computational graph and an averaged loss applied to the attention coefficients for caption generation, inspired by~\cite{XuK}.

\textbf{Contributions}. Summarising, 
to the best of our knowledge: (1) our work presents the first End-to-End trainable framework (EtENet) for video captioning, designed to learn task-specific features, based on a new two-stage efficient training strategy.
Our approach propagates the gradient from the last layer of the RNN decoder to the first layer of the CNN encoder 
as illustrated in Figure~\ref{fig:gradientflow} (B). 
Unlike~\cite{Zhang, Wang, Pan, Yao} which all use more than 25 frames per video clip, our model only uses 16 frames, a significant contribution in terms of resource economics.
\\
(2) Performance-wise, training our network architecture in the traditional, disjoint way produces comparable results to the current state-of-the-art, whereas our end-to-end training framework delivers significant performance improvements no matter the choice of the base encoder network. This \emph{sets a new benchmark for the field of video captioning}, upon which further progress can be made. Notably, this is achieved without using any 3D CNN encoding.
\\
Finally (3), inspired by~\cite{Yao}, we present and plug into our end-to-end framework an overhauled version of the Soft-Attention decoder characterised by improved performance compared to the original version.

\iffalse
Summarising, we present the first end-to-end trainable framework for video captioning which:
\begin{itemize}
    \item can learn to encode video captioning-specific features;
    \item accumulates the gradients to limit GPU memory consumption;
    \item uses a two-state training process to speed up training;
    \item establishes a effective and simple baseline for future works.
\end{itemize}
\fi

\section{Related work}

Inspired by the latest computer vision and machine translation techniques~\cite{Sutskever}, recent efforts in the video captioning field follow a \emph{sequence learning} approach. The commonly adopted architecture, as mentioned, is an encoder-decoder framework~\cite{Wu} that uses either 2D or 3D CNNs to collect video features in a fixed-dimension vector, which is then fed to a Recurrent Neural Network to generate the desired output sentence, one word at a time.

The first notable work in this area was done by Venugopalan \etal~\cite{Venugopalan}. They represented an entire video using a mean-pooled vector of all features generated by a CNN at frame level. The resulting fixed-length feature vector was then fed to an LSTM for caption generation. Although state-of-the-art results were achieved at the time, the temporal structure of the video was not well modelled in this framework. Since then, alternative views have been supported on how to improve the visual model in the encoder-decoder pipeline. The same authors~\cite{Venugopalan2} have later proposed a different method which exploits two-layer LSTMs as both encoder and decoder. In this setting, compared to the original pipeline~\cite{Venugopalan}, each frame is used as input at each time step for the encoder LSTM, which takes care of encoding the temporal structure of the video into a fixed size vector. This model, however, still leaves room for a better spatiotemporal feature representation of videos, as well as calling for improved links between the visual model and the language model.

To address this problem, 3D CNNs and attention models have been since introduced. Inspired by~\cite{XuK}, Yao \etal~\cite{Yao} employ a temporal Soft Attention model in which each output vector of the CNN encoder is weighted before contributing to each word's prediction. The spatial component is extracted using the intermediate layers of a 3D CNN which is used in combination with the 2D CNN. On their part, \cite{Yu} have proposed a spatiotemporal attention scheme which includes a paragraph generator and sentence generator. 
The paragraph generator is designed to pick up the sentence's ordering, whereas the sentence generator focuses on specific visual elements from the encoder.

A distinct line of research has been brought forward by Pam and his team in~\cite{Pan} and subsequently in~\cite{Pan2}. 
The first work tries, in addition to using features from both 2D and 3D CNNs, to introduce a visual semantic embedding space for enforcing the relationship between the semantics of the entire sentence and the corresponding visual content. 
Multiple Instance Learning models have been used in~\cite{Fang} for detecting attributes to feed to a two-layer LSTM controlled by a transfer unit.

In the last two years, numerous relevant papers have been published. Similarly to ~\cite{Pan2}, Zhang \etal~\cite{Zhang} use a task-driven dynamic fusion across the LSTM to process the different data types. The model adaptively chooses different fusion patterns according to task status. Xu \etal~\cite{XuJ2} test on the MSR-VTT dataset a Multimodal Attention LSTM Network that fuses audio and video features before feeding the result to an LSTM multi-level attention mechanism. In~\cite{Gan}, the authors create a Semantic Compositional Network plugging into standard LSTM decoding the probabilities of tags extracted the frames, in addition to the usual video features, merged in a fixed-dimension vector. Chen \etal~\cite{ChenY} show that it is possible to get good results using just $\sim$6-8 frames. An LSTM encoder takes a sequence of visual features while a GRU decoder helps generate the sentence. The main idea of this interesting work is to use reinforcement learning, while a CNN is used to discriminate whether a frame must be encoded or not. A recent work~\cite{Wang} improves the performance of~\cite{Yao} and its model architecture by inserting a reconstruction layer on top that aims to replicate the video features, starting from the hidden state of the LSTM cell.

Unlike all previous work, in which the weights of the encoder part do not change, our study focusses on end-to-end training. This strategy, which has been proven successful in various applications, encourages the encoder to capture features which are actually discriminant for caption generation. Our training process is divided into two stages: while in the first stage only the decoder is trained, in the second one the whole model is fine-tuned. In this work, in particular, we test two different 2D CNN  encoders: GoogLeNet~\cite{Szegedy} and Inception-ResNet-v2~\cite{Szegedy2}, thus showing how end-to-end training is beneficial no matter the choice of the base encoder. This also demonstrates that our training strategy is not linked to any particular encoder network. As decoder, we present a modified version of the Soft-Attention model defined by~\cite{Yao}. Nevertheless, the approach is entirely general and can be applied to other decoding architectures.
\section{Approach}\label{approach}

\begin{figure}[t]
\begin{center}
%\fbox{\includegraphics[width=0.8\linewidth]{img/fig2.png}}
\includegraphics[width=1\linewidth]{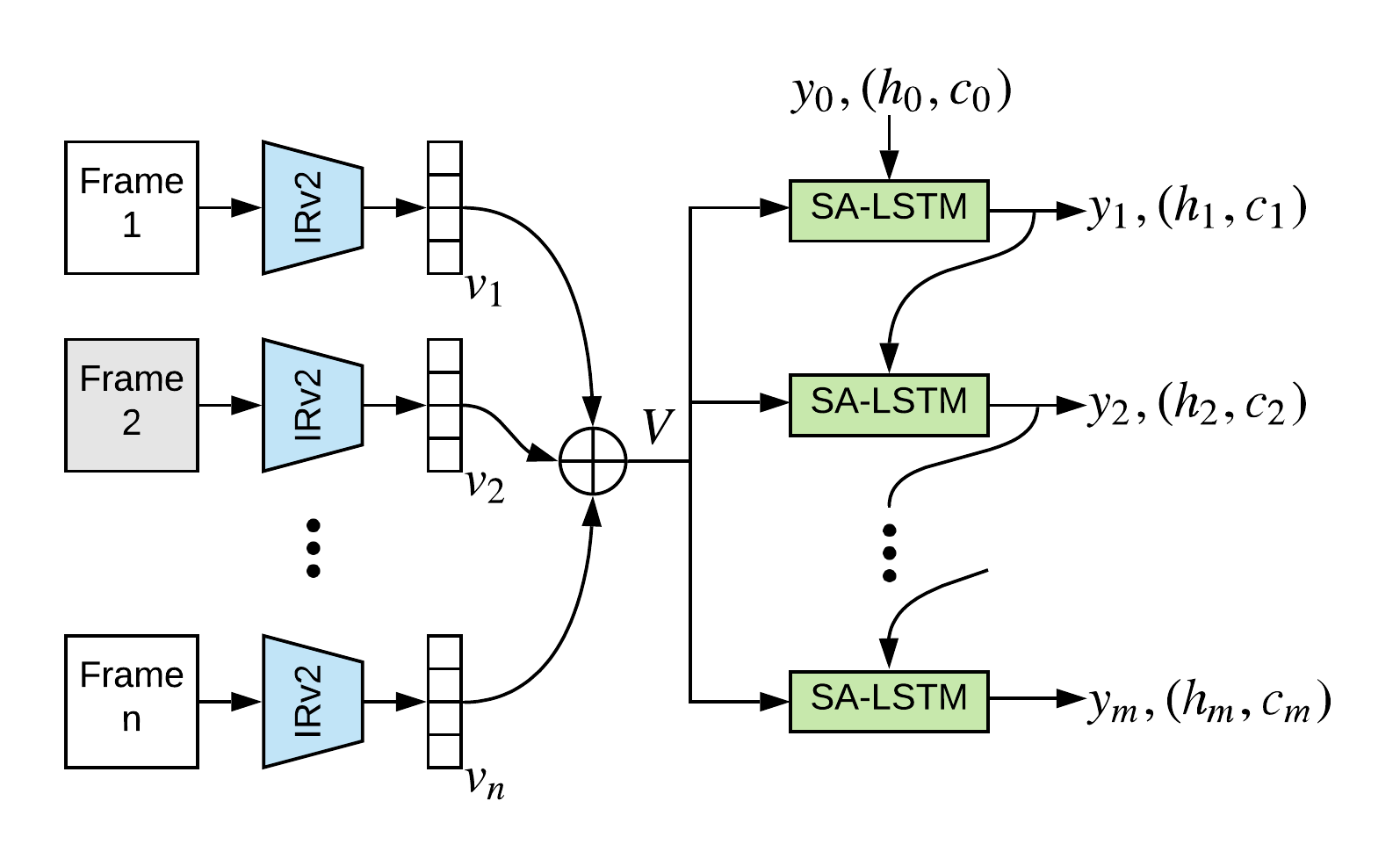}
\end{center}
\caption{Diagram of our EtENet-IRv2 architecture. Each frame is processed individually using Inception-ResNet-v2 (IRv2). All resulting vectors $v_1, ..., v_n$ are concatenated ($\oplus$) to represent their collection $V$. At each decoding time step $t$ a single word is predicted by SA-LSTM, based on the vector $V$, the previous word $y_{t-1}$ and the hidden states $(h_{t-1}, c_{t-1})$. \vspace{-3mm}}
\label{fig:model}
\end{figure}

This section describes in detail our end-to-end trainable encoder-decoder architecture for video-captioning.
The overall framework is depicted in Figure \ref{fig:model}. Section~\ref{subsec:encoder} illustrates the chosen encoder, based on the Inception-ResNet-v2~\cite{Szegedy2} architecture. The decoder, based on our original Soft-Attention LSTM (SA-LSTM) design inspired by~\cite{Yao}, is discussed in \S~\ref{subsec:decoder}. Difference with~\cite{Yao} %in \S~\ref{subsubsec:improvements} 
and initialisation details are also discussed there. %in \S~\ref{subsubsec:init}. 
Note that our framework is also tested using GoogLeNet~\cite{Szegedy} as the encoder.

The training process is explained in \S~\ref{subsec:training}. Firstly, gradient accumulation %(\S~\ref{subsubsec:gradaccum})
is used to achieve the desired high batch size for the training of the decoder. 
Secondly, the proposed two-stage training process is described. %to speed up the training process in \S~\ref{subsec:training}. 
Lastly, the proposed averaged loss functions are defined and justified. % in \S~\ref{subsubsec:loosfun}.

\subsection{Encoder}\label{subsec:encoder}

A common strategy~\cite{Venugopalan} is to use a 2D CNN pre-trained on the ImageNet dataset~\cite{Russakovsky} as an encoder.
Typically, feature vectors are generated before the first fully-connected layer of the neural network, for each frame of the video. 
This is done as a preprocessing step, and many versions of 2D CNN were brought forward over the course of the years for video captioning. For instance, Venugopalan \etal~\cite{Venugopalan, Venugopalan2} would use variants of AlexNet~\cite{Krizhevsky}, 16 layer VGG~\cite{Simonyan} and GoogLeNet~\cite{Szegedy}. Yao \etal~\cite{Yao} would also use  GoogLeNet, in combination with a 3D CNN. Gan \etal~\cite{Gan}, instead, preferred ResNet-152~\cite{He} whereas Wang \etal~\cite{Wang} used Inception-v4~\cite{Szegedy2}. 

As noted by Wang~\etal~\cite{Wang}, deeper networks are more likely to capture the high-level semantic information about the video. Thus, in this work we decided to use Inception-ResNet-v2~\cite{Szegedy2} as the encoder, among all possible convolutional network architectures. 
The version used in our experiments is pre-trained on ImageNet, and is publicly available\footnote {\url{https://github.com/Cadene/pre-trained-models.pytorch}}.
%in the \verb'Cadene/pre-trained-models.pytorch' repository
To show the generality of our framework,
additional tests were conducted using our own PyTorch version of GoogLeNet~\cite{Szegedy}, also pre-trained on ImageNet. The weights were imported from a well known Caffe version\footnote{\url{https://github.com/BVLC/caffe/tree/master/models/bvlc_googlenet}}.

Formally, given a video $X = \{x_1, ..., x_n\}$ composed by a sequence of $n$ RGB images, 
our encoder $\phi$ is a function mapping each frame $x_i$ to a feature vector $v_i$ using the average pooling stage after the \verb'conv2d_7b' layer of Inception-ResNet-v2 or, in the case of GoogLeNet, the \verb'pool5/7x7 s1' layer. 
We thus denote by
\begin{equation}
V = \{v_1, ..., v_n\} = \phi(\{x_1, ..., x_n\})
\end{equation}
the output of the encoder, later fed as input to the decoder.

\subsection{Decoder: Soft Attention LSTM}~\label{subsec:decoder}

Using Inception-v4 as decoder,
%The work by Wang \etal~\cite{Wang} was instrumental to the performance exhibited by
the Soft-Attention model developed by Yao \etal~\cite{Yao} achieves good performance compared to other, more complex systems (e.g.,~\cite{Yu}).
\\
For this reason our work builds on the version of \cite{Yao} developed in Theano\footnote{\url{https://github.com/yaoli/arctic-capgen-vid}}. 
%For ease of implementation
%With the advent of new frameworks such as TensorFlow and PyTorch, working with Theano as a base of our work would have been quite difficult. Thus, 
%we decided to create our own version of the decoder proposed in \cite{Yao} written using PyTorch.
The resulting original SA-LSTM framework, implemented in PyTorch, exhibits a number of significant improvements over \cite{Yao},
%\\ Our SA-LSTM framework contemplates a number of variants to the original formulation, which are 
explained in detail below.

\textbf{Formulation}.
At each decoding step the SA-LSTM $\psi$ takes as input $n$ vectors generated by the encoder, $V = \{ v_1, ..., v_n \}$, together with the previous hidden state $h_{t-1}$, memory cell $c_{t-1}$ and word $y_{t-1}$. Its output is composed by: (i) the probability of the next word $P(y_t|y_{<t}, V)$ based on the previously observed words $y_{<t}$ and on the feature vectors $V$; (ii) 
the current hidden state $h_t$, and (iii) the current memory cell state $c_t$.
Namely:

\begin{equation}
\left[
  \begin{tabular}{c}
  $P(y_t|y_{<t}, V)$ \\
  $h_t$ \\
  $c_t$ 
  \end{tabular}
\right] 
= \psi(V, y_{t-1}, h_{t-1}, c_{t-1}).
\end{equation}
The algorithm runs sequentially through the output sequence, predicting one word at a time.

More in detail, at any given time $t$ the first step is to create a single vector from $V$ by applying a Soft-Attention mechanism $\varphi_t$ to the whole encoder output. Firstly, for each vector $v_i$ a normalised score $\alpha_i^{t}$ is computed:
\begin{equation}
e_i^{t} = W_a^\top \tanh(W_{eh} h_{t-1} + W_{ev} v_i + b_e) + b_a;
\end{equation}
\begin{equation}
\alpha_i^{t} = \frac{\exp\{e_i^{t}\}}{\sum_{j}^{\mid V \mid}\exp\{e_j^{t}\},}
\end{equation}
where $W_a$, $W_{eh}$, $W_{ev}$, $b_e$, $b_a$ are all trainable variables and $e_i^{t}$ is the unnormalised score of vector $i$ at time $t$.
\\
Secondly, a coefficient $\beta_t\in[0,1]$ is computed to measure the importance of the final vector $\varphi_t(V)$ as a function of the previous hidden state, with parameters $W_\beta$ and $b_\beta$:
\begin{equation} \label{sa-lstm_beta}
\beta_t = \sigma(W_\beta h_{t-1} + b_\beta),
\end{equation}
where $\sigma$ is a sigmoid activation function.
The final output of the Soft-Attention function is computed as follows:
\begin{equation} \label{eq:final-vector}
\varphi_t(V) = \beta_t \sum_{i}^{\mid V \mid} \alpha_i^{t} v_i.
\end{equation}
The vector (\ref{eq:final-vector}) 
%of the Soft-Attention function 
is then concatenated with an embedding $E$ of the previous word $y_{t-1}$
\begin{equation} \label{lstm_input}
z_t = [\varphi_t(V), E[y_{t-1}]],
\end{equation}
and fed to a standard LSTM with state vector $h_t$:
\iffalse
Memory cell and hidden state are updated through the following equations:
\begin{equation} \label{lstm_init}
i_t = \sigma(W_{ii} z_t + b_{ii} + W_{hi} h_{t-1} + b_{hi});
\end{equation}
\begin{equation}
f_t = \sigma(W_{if} z_t + b_{if} + W_{hf} h_{t-1} + b_{hf});
\end{equation}
\begin{equation}
g_t = \tanh(W_{ig} z_t + b_{ig} + W_{hg} h_{t-1} + b_{hg});
\end{equation}
\begin{equation}
o_t = \sigma(W_{io} z_t + b_{io} + W_{ho} h_{t-1} + b_{ho});
\end{equation}
\begin{equation}
c_t = f_t * c_{t-1} + i_t * g_t;
\end{equation}
\begin{equation} \label{lstm_end}
h_t = o_t * \tanh(c_t).
\end{equation}
\fi
\begin{equation} \label{lstm_end}
(h_t, c_t) = LSTM(z_t, h_{t-1}, c_{t-1}).
\end{equation}
The word prediction $p_t = P(y_t|y_{<t}, V)$ is a function of the concatenation of $\varphi_t(V)$ and $h_t$, and of the embedding $E[y_{t-1}]$ of $y_{t-1}$:
\begin{equation} \label{sa-lstm_u}
u_t = W_u [\varphi_t(V), h_t] + E[y_{t-1}] + b_u;
\end{equation}
\begin{equation}
p_t = softmax(W_p \tanh(u_t) + b_p),
\end{equation}
where $[\cdot,\cdot]$ denotes vector concatenation and all weight matrices $W$ and bias vectors $b$ are trainable parameters.

\textbf{Decoder innovations}. %~\label{subsubsec:improvements}
Our decoder architecture differs from \cite{Yao}'s in a number of ways.
%We need to stress a number of differences with the original framework as formulated in~\cite{Yao}: 
(i) In Equation (\ref{sa-lstm_u}), $E[y_{t-1}]$ is not mapped by a matrix parameter. The effect of this is that the  visual information adapts to the word embedding, rather than the opposite, as in residual connection frameworks \cite{He}.
(ii) In Equation (\ref{sa-lstm_beta}), a $\beta_t$ term is added to reflect the fact that for some connective words in the sentence (e.g. 'the') the attention term should weigh less.
(iii) An averaged Doubly Stochastic Attention is used as attention loss (as discussed in Sec. \ref{subsubsec:loosfun}), for this has shown to improve performance.
%FAB: we need to explain the reasons why
Implementation-wise, (iv) the LSTM machinery (Equations (\ref{lstm_input}) to (\ref{lstm_end})) is derived from
\verb'torch.nn.LSTMCell', rather than having been implemented from scratch.

%These changes are inspired by the original code repository by Yao \etal. After publication of~\cite{Yao}, the authors released new tests including the above suggestions for model improvement.
Another substantial difference between the two models lies in their computational graphs at training time. In \cite{Yao} Soft-Attention is a \emph{sentence-by-sentence} model, meaning that the dimensionality of the decoder input is (batch\_size, sequence\_length, embedding\_dim+encoder\_output\_dim).
In our decoder we use a \emph{word-by-word approach} with as input shape: (batch\_size, embedding\_dim+encoder\_output\_dim). 
%The main difference between these two approaches is that, 
\\
The latter first computes the entire sequence of the LSTM hidden states $\{h_t\}$ to then produce a sentence.  
The former generates one hidden state vector at a time, and the word is directly predicted at every step. This difference affects the gradient upgrading procedure during back-propagation. 

\textbf{Initialization Details}. %~\label{subsubsec:init}
Our framework takes blocks of 16 RGB frames as input. Each frame is processed by our version of Inception-ResNet-v2 up to the average pooling stage after the \verb'conv2d_7b' layer. Thus, the encoder output $V$ is composed by 16 vectors of 1536 elements each.
In the GoogLeNet case we use the output of the
\verb'pool5/7x7 s1' 
layer, so that the vector has 1024 elements.
\\
As explained, at each step the decoder takes as input $V$, the previously observed word $y_{t-1}$ and the hidden $h_{t-1}$ and $c_{t-1}$ states. The first word of every predicted sentence is the token \verb'<SOS>', while $h_0$ and $c_0$ are initialised as follows:
\begin{equation}
h_0 = \tanh(W_h \overline{V} + b_h);
\end{equation}
\begin{equation}
c_0 = \tanh(W_c \overline{V} + b_c),
\end{equation}
where $W_h$, $W_c$, $b_h$, $b_c$, are trainable variables and $\overline{V}$ is the mean of all the vectors in $V$.
\\
The desired video caption is predicted word by word until  \verb'<EOS>' is produced or after a maximum caption length is reached (set to 30 for MSVD and to 20 for MSR-VTT). The input $V$ is the same throughout each iteration.
We use 512 as the dimension of the LSTM hidden layer, 486 as embedding dimension for $E[y_{t-1}]$, while the cardinality of the word probability vector $p_t$ obviously depends on the size of the vocabulary being considered ($\sim$12,000 for MSVD, $\sim$200,000 for MSR-VTT).

\subsection{Training Process}\label{subsec:training}

\textbf{Accumulate to Optimize}. \label{subsubsec:gradaccum}
Recurrent networks require a large (e.g. 64 in ~\cite{Yao}) batch size to converge to good local minima. 
This is true for our SA-LSTM as well, since it is based on an LSTM recurrent architecture.
\\
In our initial tests, when using a disjoint training setup similar to~\cite{Yao}'s, we noticed that increasing the batch size would indeed boost performance.
Unfortunately, Inception-ResNet-v2 (as other CNNs) is very expensive in term of memory requirements, hence large batch sizes are difficult to implement. A single batch, for instance, would use 5 GB (GigaByte) of GPU memory. The machine our tests were conducted on comes with 4 Nvidia P100 GPUs with 16 GB of memory each, allowing a maximum batch size of 12. 

To overcome this problem, our training strategy is centred on accumulating gradients until the neural network has processed 512 examples. After that, the accumulated gradients are used to update the parameters of both encoder and decoder. The pseudocode for this process is provided in Algorithm \ref{alg:accumulate_gradient}. The standard training process is modified into one that accumulates gradients for $accumulate\_step$ size. As a result, the approach achieves an effective batch size equal to $accumulate\_step \times mini\_batch\_size$.

\begin{algorithm}
\caption{Training with accumulated gradient. \label{alg:accumulate_gradient}}
\begin{algorithmic}[1]
\REQUIRE $accumulate\_step$
\STATE $i \leftarrow 0$
\STATE Reset gradient to zero
\FOR{batch size of Examples in Training set}
\STATE Model forward step using Examples
\STATE Compute loss
\STATE Normalise loss using $accumulate\_step$
\STATE Backward step and accumulate gradients
\IF{$(i \mod accumulate\_step)$ is $0$}
\STATE Update model with accumulated gradient
\STATE Reset gradient to zero
\ENDIF
\STATE $i \leftarrow i + 1$
\ENDFOR
\end{algorithmic} 
\end{algorithm} 

\textbf{Two-Stage Training}. \label{subsubsec:twostage}
Stochastic optimisers require many parameter update iterations to identify a good local minimum. Hence, if naively implemented, our gradient accumulation strategy would be quite slow, as opposed to disjoint training in which GPU memory requirements are much lower. To strike a balance between a closer to optimal but slower end-to-end training setup and a faster but less optimal disjoint training setting we adopt a \emph{two-stage training} process, which is also crucial to allow end-to-end training of an heterogenous network from a computational standpoint.

\emph{In the first stage}, we freeze the weights of the pre-trained encoder to train the decoder. As the encoder's weights are kept constant, this is equivalent to training a decoder on pre-computed features from the encoder.
As a result, memory requirements are low, and the process is fast. Once the decoder reaches a reasonable performance on the validation set, the second stage of the training process starts. 
\\
\emph{In the second stage}, the whole network is trained end-to-end while freezing the batch normalisation layer (if the encoder contemplates it). 

In both stages, at each time step SA-LSTM uses the real target output (i.e., the target word) as  input, rather than its own previous prediction.
\\
Given the heterogeneity of the architecture,
we use Adam~\cite{Hochreiter} as an optimisation algorithm and different parameter values for encoder and decoder. 
For the former, inspired by~\cite{Szegedy, Szegedy2}, since the batch size is 512 and each example has 16 frames, we set the learning rate to $1e-09$ and the weight decay to $4*1e-05$ in the experiments with Inception-ResNet-v2. The version with GoogLeNet uses a learning rate of $2e-04$ and a weight decay of $2*1e-04$. 
The decoder is instead updated using $1e-04$ as learning rate and $1e-04$ as weight decay. To avoid the vanishing and exploding gradient problems typical of RNNs, we force the gradient to belong to the range $[-10,10]$.
At test time a beam search~\cite{Norvig} with size 5 is used for final caption generation as in~\cite{Wang}.
Larger sizes do not improve performance.

\textbf{Loss Function}. \label{subsubsec:loosfun}
Similarly to ~\cite{Yao,XuK}, we adopted as overall loss of the network the following sum:
\begin{equation}
\mathcal{L}_{tot}(\theta) = \mathcal{L}(\theta)_{NLL} + \lambda \mathcal{L}(\theta)_{aDSA}.
\end{equation}
The Negative Log-Likelihood loss is given by:
$
\mathcal{L}(\theta)_{NLL} = - \sum_{i}^{N} \sum_{t}^{C} \log p(y_t^i | y_{<t}^i, x_i, \theta),
$
where $C$ is the caption length, $N$ the number of examples. 
The second component is an original \emph{averaged 
Doubly Stochastic Attention} loss:
\begin{equation} \label{dsa-loss}
\mathcal{L}(\theta)_{aDSA} = \frac{1}{L} \sum_{k}^{L} \left (1 - \sum_{t}^{C} \alpha_k^t \right )^2,
\end{equation}
which we found improves performance over the standard DSA one,
with $L$ the size of the feature vector $v_i$. 
The (average) DSA component of the loss can be seen as encouraging the model to pay equal attention to every frame over the course of the caption's generation process.

Similarly to ~\cite{Yao}, we set $\lambda$ to $0.70602$.

\section{Evaluation}

Before discussing our tests, we first describe the metrics (\S~\ref{metrics}) and datasets (\S~\ref{subsec:datasets}) used to evaluate our model, and the pre-processing steps applied to the input data (\S~\ref{subsec:preprep}). 

\subsection{Metrics} \label{metrics}
To guarantee a fair quantitative comparison with the state of the art we used the most common and well known metrics: BLEU~\cite{Papineni} (4-gram version), METEOR~\cite{Banerjee}, ROUGE-L~\cite{Lin} and CIDEr~\cite{Vedantam}.
While BLEU and METEOR were created for machine translation tasks, ROUGE-L's aim is to compare a machine-generated summary with the human-generated sentence. CIDEr is notable as  the only metric created for evaluating image descriptions that use human consensus.

\subsection{Datasets}\label{subsec:datasets}

We evaluated our model and compared it with our competitors on two standard video captioning benchmarks.
MSVD is one of the first such datasets to include multi-category videos. MSR-VTT, on its side, is based on 20 categories and is of a much larger scale than MSVD.

{\bf MSVD.} The most popular dataset for video captioning systems evaluation is, arguably, the Microsoft Video Description Corpus (MSVD), also known as YoutubeClips~\cite{ChenDL}. The dataset contains 1970 videos, each video depicting a single activity lasting about 6 to 25 seconds. 
%The frame rate is in most cases 30 frames per second. 
Each video is associated with multiple descriptions in the English language, collected via Amazon Mechanical Turk, for a total of 70,028 natural language captions. We adopted the evaluation setup of ~\cite{Venugopalan}, and
splitted the dataset into three parts: 1,200 videos for training, 100 videos for validation, and the remaining 670 videos for testing.

{\bf MSR-VTT.} The MSR Video to Text~\cite{XuJ} dataset is a recent large-scale benchmark for video captioning. 10K video clips from 20 categories were collected from a commercial video search engine (e.g., music, sports, and TV shows). Each of these videos was annotated with 20 sentences produced by 1327 Amazon Mechanical Turk workers, for a total number of captions of around 200K. 
\\
As prescribed in the original paper, we split videos by index number: 6,513 for training, 497 for validation and 2,990 for the test. The number of unique words present in the captions is close to 30K.

\subsection{Preprocessing}\label{subsec:preprep}

Following the usual pre-processing of Inception-ResNet-v2, height and width of each frame of the video were resized to 314, to then use the central crop patch of 299x299 pixels of each frame. Pixel normalisation using a mean and standard deviation of 0.5 was applied. Training, validation and test examples were all subject to the same frame preprocessing steps. 
%To save memory space, the images are saved as u-int8 while pixel normalisation is performed at each time instant. 
For the GoogLeNet encoder, we used the commonly accepted preprocessing steps for that network: each frame of the video was resized to 224x224 pixels, and then normalised by mean subtraction.
%for every channel: R$=123.68$, G$=116.779$, B$=103.939$. The channel input of the network is BGR.
\\
Using all frames of a video is very time inefficient -- as~\cite{ChenY} shows, it is possible to create an efficient model using fewer frames. On the other hand, we did not apply any additional filtering to the frames, as we preferred to leave this task for the attention mechanism to handle. 
In agreement with~\cite{Wang} and with our findings, we decided to represent each video by 16 equally-spaced features.

As for the captions, we tokenised them by converting all words to lowercase and applying the \verb'TreebankWordTokenizer' class from the Natural Language Toolkit\footnote{\url{https://www.nltk.org}} to split sentences into tokens. The tokeniser uses regular expressions as in the Penn Treebank\footnote{\url{http://www.cis.upenn.edu/~treebank/tokenizer.sed}}, thus adhering to English grammar while maintaining punctuation in the token.

\section{Experiments}\label{sec:experiments}

We conclude by reporting and discussing the experimental validation of our end-to-end trainable framework (EtENet) on the datasets described in (\S~\ref{subsec:datasets}).

\subsection{SA-LSTM vs Soft Attention}

As a first step, we compared the performance of our SA-LSTM$_{base}$ decoder (in PyTorch) with that of Soft Attention~\cite{Yao} (Theano), with the same parameter values: $learning\_rate=0.01$, $batch\_size=64$, Adadelta as optimizer, and using the original DSA loss~\cite{XuK} rather than our bespoke, averaged version. %FAB: we did not really explained that in 3.3
The version of our decoder we term
SA-LSTM$_{best}$, instead, improves on SA-LSTM$_{base}$ by using different hyperparameter values, optimisation algorithm and loss. Namely, the best results are achieved using $learning\_rate=0.0001$, $batch\_size=512$, Adam as optimizer and our averaged DSA loss (\ref{dsa-loss}).
Note that the original approach requires a lot of memory, so that the higher batch size possible on our machines was $64$. Using our model, instead, we could achieve a value of $512$. Note also that we only used one GPU for training for sake of fair comparison, as Theano does not support multi-GPU training. The results are shown in Table~\ref{tab:originalvsour}. 
\\
For this comparison we used the features extracted from GoogLeNet stored in the original SA repository~\cite{Yao}. 
%that it is the same of our version of the GoogLeNet, 
Thus, the results are not comparable with those of Table~\ref{tab:MSVD}.

\begin{table}[h!]
\begin{center}
\begin{tabular}{|l|cccc|}
\hline
Model & B@4 & M & C & R-L \\
\hline\hline
SA & 46.6 & 32.0 & 67.0 & 68.0 \\
SA-LSTM$_{base}$ & 46.9 & 32.1 & 70.9 & 69.2 \\
SA-LSTM$_{best}$ & 48.3 & 32.2 & 76.4 & 69.1 \\
\hline
\end{tabular}
\end{center} \vspace{-2mm}
\caption{Comparison between the original Soft-Attention (SA) decoder and ours (SA-LSTM), on the test set of the MSVD dataset.}
\label{tab:originalvsour}
\end{table}

Both versions of our decoder outperform \cite{Yao} -- by a very significant amount in the optimised (best) version, especially in the CIDEr metric.

\subsection{State-of-the-art Comparison}\label{subsec:results}

Tables \ref{tab:MSVD} and \ref{tab:MSR-VTT} clearly show how our approach, both when using only step 1 of the training, and when applying both steps, matches or outperforms all the work done previously using  Inception-ResNet-v2 as the encoder (EtENet-IRv2), except when measured using the BLEU metric.
In fact, as explained by Banerjee~\etal~\cite{Banerjee}, BLEU is a metric that has many weaknesses, e.g., the lack of explicit word-matching between translation and reference. 
\\
In opposition, according to \cite{Vedantam}, CIDEr was specifically designed to evaluate automatic caption generation from visual sources, and is thus arguably more relevant. 
%Hence, it makes more sense to stress the results under the CIDEr metric. 
Indeed, our proposed EtENet-IRv2 outperforms all the existing state-of-the-art method across both datasets (see Tables~\ref{tab:MSVD},~\ref{tab:MSR-VTT}) when performance is measured by CIDEr.

\subsection{Discussion}

The substantial difference
between our model and the others assessed confirms that EtENet-IRv2 succeeds in achieving excellent results without requiring an overly complex structure, e.g., the addition of new layers as in RecNet (row 11, Table~\ref{tab:MSVD}), or the adoption of new learning mechanisms such as reinforcement learning as in PickNet (row 3, Table~\ref{tab:MSR-VTT}). Moreover, this shows that it is possible to obtain excellent results even when using roughly half the frames used in other competing approaches~\cite{Yao,XuJ2, Zhang, Wang}.
\emph{Our framework sets a new standard in terms of top performances in video captioning} and, we believe, can much contribute to 
%which deserve to be disseminate for the sake of 
further progress in the field. Additionally, this is done without resorting to fancy 3D CNN architectures, thus leaving huge scope for further improvements. Moreover, unlike~\cite{Zhang, Wang, Pan, Yao} which all use more than 25 frames per video clip, our model only uses 16 frames, a significant contribution in terms of memory and computational cost.

The performance of EtENet-GLN, which uses GoogLeNet as encoder, is comparable to that of all mechanism using similar or older versions of the decoder (e.g., VGG)~\cite{Venugopalan, Venugopalan2, Yao, Pan, Yu, Pan2, XuJ2}. Notably, though, it does achieve its best results in the CIDEr metric.

% MSVD Dataset
\begin{table}[t!]
\begin{center}
\begin{tabular}{|l|cccc|}
\hline
Model & B@4 & M & C & R-L \\
\hline\hline
LSTM-YT~\cite{Venugopalan} & 33.3 & 29.1 & - & - \\
S2VT~\cite{Venugopalan2} & - & 29.8 & - & - \\
SA~\cite{Yao} & 41.9 & 29.6 & 51.7 & - \\
LSTM-E~\cite{Pan} & 45.3 & 31.6 & - & - \\
h-RNN~\cite{Yu} & 49.9 & 32.6 & 65.8 & - \\
LSTM-TSA~\cite{Pan2} & 52.8 & 33.5 & 74.0 & - \\
SCN-LSTM~\cite{Gan} & 51.1 & 33.5 & 77.7 & - \\
MA-LSTM~\cite{XuJ2} & 52.3 & 33.6 & 70.4 & - \\
TDDF~\cite{Zhang} & 45.8 & 33.3 & 73.0 & 69.7 \\
PickNet~\cite{ChenY} & {\bf52.3} & 33.3 & 76.5 & 69.6 \\
RecNet~\cite{Wang} & {\bf52.3} & 34.1 & 80.3 & 69.8 \\
\hline
EtENet-GLN$_{step1}$ & 48.2 & 32.0 & 75.1 & 68.9 \\
EtENet-GLN$_{step2}$ & 48.9 & 32.4 & 78.0 & 69.4 \\
\hline
EtENet-IRv2$_{step1}$ & 49.1 & 33.6 & 83.5 & 69.5 \\
EtENet-IRv2$_{step2}$ & 50.0 & {\bf34.3} & {\bf86.6} & {\bf70.2} \\
\hline
\end{tabular}
\end{center}
\caption{Comparison between our architecture (EtENet), using both GoogLeNet (GLN) and IRv2 as the encoder, and the state-of-the-art on the MSVD dataset.}
%using the following metrics: BLEU-4, METEOR, CIDEr and ROUGE-L.}
\label{tab:MSVD}
\end{table}

\subsection{Impact of End-to-End Training}

%FAB: non dovevamo mostrare i risultati del training end to end from stratch (cioe' senza preinizializzatione)? Questo mi sembra diverso dal mostrare l'effetto di step 1+step 2

Importantly, for both incarnations of our architecture (EtENet-GLN e EtENet-IRv2) end-to-end training (step 2) positively impact the performance across the board (i.e., across datasets and metrics), thanks to the additional fine tuning. Our network is able to match or outperform the other state-of-the-art models, while very significantly outperforming the Soft-Attention (SA) approach \cite{Yao} (third row of Table~\ref{tab:MSVD}).

% MSR-VTT Dataset
\begin{table}[h!]
\begin{center}
\begin{tabular}{|l|cccc|}
\hline
Model & B@4 & M & C & R-L \\
\hline\hline
MA-LSTM~\cite{XuJ2} & 36.5 & 26.5 & 41.0 & 59.8 \\
TDDF~\cite{Zhang} & 37.3 & {\bf27.8} & 43.8 & 59.2 \\
PickNet~\cite{ChenY} & {\bf41.3} & 27.7 & 44.1 & 59.8 \\
RecNet~\cite{Wang} & 39.1 & 26.6 & 42.7 & 59.3 \\
\hline
EtENet-IRv2$_{step1}$ & 40.3 & 27.5 & 46.8 & 60.4 \\
EtENet-IRv2$_{step2}$ & 40.5 & 27.7 & {\bf47.6} & {\bf60.6} \\
\hline
\end{tabular}
\end{center}
\caption{Comparison between our EtENet-IRv2 architecture and the state-of-the-art on the MSR-VTT benchmark, using the following metrics: BLEU-4, METEOR, CIDEr and ROUGE-L. \vspace{-3mm}}
\label{tab:MSR-VTT}
\end{table}

\subsection{Qualitative results}

\begin{figure}[t]
\begin{center}
%\fbox{\includegraphics[width=0.85\linewidth]{img/fig1.png}}
\includegraphics[width=0.95\linewidth]{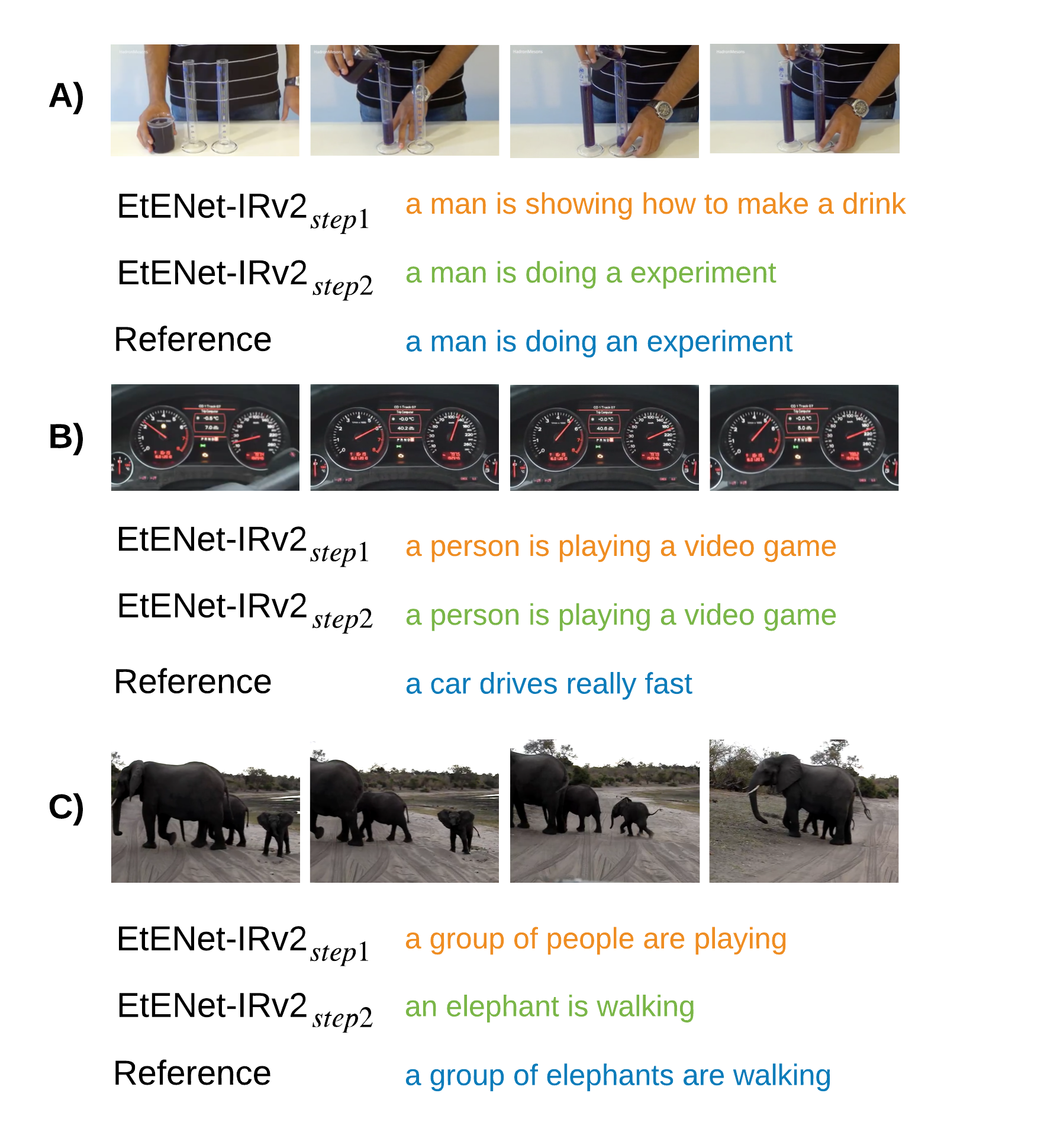}
\end{center}
\caption{Qualitative results produced by EtENet-IRv2. In (A) and (C), which show a video from MSR-VTT and one from MSVD, respectively, it is possible to observe how end-to-end training can dramatically improve the quality of the resulting caption. In (B) a negative example from the MSR-VTT dataset is shown, for which our network cannot successfully identify the ground truth.}
\label{fig:qualitative}
\end{figure}

From a qualitative point of view, Figure  \ref{fig:qualitative} reports both some positive and some negative examples. Generally, we can notice that the increase in accuracy achieved by the two-step training setting leads, in some cases, to a visible improvement of the generated sentences.

Much more extensive quantitative results are reported in the Supplementary Material.

\section{Conclusions} \label{sec:conclusions}

In this paper, we proposed a simple end-to-end framework for video-captioning. 
To address the problem with the large amount of memory required to process video data for each batch, a gradient accumulation strategy was conceived. We proposed a training procedure articulated into two steps to speed up the training process, and allow efficient end-to-end training. 
Our evaluation on standard benchmark datasets showed how our approach outperforms the state of the art using all the most commonly accepted metrics.
\\
We believe we managed to set a new baseline for future work thanks to our principled end-to-end architecture, providing an opportunity to take research in the field forward starting from a more efficient training framework. 

Our model is not exempt from {drawbacks}. Training a very deep a neural network end-to-end requires significant computational resources. Our proposed two-stage training process is a step towards an efficient training procedure suited to the task.
Further research directions include the integration of a more formal treatment of language semantics in the model, and spatio-temporal attention mechanisms based on state of the art action and object tube detectors.
%%%%%%%%%%%%%%%%%%%%%%%%%%%%%%%%%%%%%%%%%%%%%%
{\small
\bibliographystyle{ieee_fullname}
\bibliography{mainbib}
}

%\clearpage
%\input{text/suppmat.tex}

\end{document}